\documentclass[10pt,leqno]{amsart}

\usepackage{graphicx}
\usepackage{indentfirst}
\usepackage{csquotes}
\usepackage{amssymb, amsthm, amsmath}
\usepackage{xcolor}
\usepackage{paralist}
\usepackage{booktabs}

\usepackage{fancyhdr}
\usepackage{etoolbox}
\usepackage{hyperref}
\usepackage{tikz}

\hypersetup{
    colorlinks=true,
    linkcolor=black,
    urlcolor=blue,
    citecolor=black,
    pdfborder={0 0 0}
}

\begin{document}

\title{Adaptive Focus Memory for Language Models}
\author[]{Christopher Cruz}
\date{\today}
\address{Purdue University}
\email{cchris2004@gmail.com, cruz209@purdue.edu}

\maketitle

\let\thefootnote\relax

\begin{abstract}
Large language models (LLMs) are increasingly deployed in multi-turn dialogue
settings, yet their behavior remains bottlenecked by naive history-management
strategies. Replaying the full conversation at every turn is simple but costly,
while recency-based truncation or static summarization often causes early,
high-impact user constraints to drift out of effective context. As a result,
models may retain text without reliably applying it when it matters.

We present \emph{Adaptive Focus Memory} (AFM), a lightweight context management
system that dynamically assigns each past message one of three fidelity
levels---\textsc{Full}, \textsc{Compressed}, or \textsc{Placeholder}---based on
semantic relevance, temporal decay, and importance classification. AFM packs
messages chronologically under a fixed token budget, preserving critical
constraints at high fidelity while allowing low-importance context to degrade
gracefully.

We evaluate AFM on two multi-turn dialogue benchmarks designed to stress
long-horizon constraint preservation: a safety-critical travel scenario
involving a user with a severe peanut allergy, and a policy-critical tax
compliance scenario involving an illegal evasion request. Under strict grading
that requires both explicit constraint recall and appropriately conditioned
generation, AFM succeeds in 83.3 percent of allergy runs where all baseline strategies fail under
strict grading, and preserves correct refusal behavior on the tax benchmark.

These results demonstrate that effective dialogue memory requires more than
retaining prior text. Selectively allocating fidelity across past messages
enables reliable constraint preservation under bounded context growth, without
modifying model weights or introducing external retrieval infrastructure. We
release an open-source implementation of AFM compatible with OpenAI-style chat
APIs to support reproducible research and practical deployment.
\end{abstract}

\bigskip

\section{Introduction}

Large language models (LLMs) such as ChatGPT, Claude, and Gemini have become
powerful general-purpose dialogue agents. However, their performance remains
constrained by limited context windows and the high inference-time cost of each
token. As conversations grow in length, na\"ively replaying the full history at
every turn leads to bloated prompts, rising latency, and mounting API
expenses---posing serious challenges for multi-agent systems, assistants, and
long-form interactions.

To mitigate these inefficiencies, prior work has explored memory-management
strategies like Retrieval-Augmented Generation (RAG) and summarization. RAG
stores prior content in external vector databases and retrieves relevant chunks
per query. Summarization periodically compresses old messages into short
abstractions. However, both approaches face limitations in dialogue settings:
RAG pipelines require dedicated retrieval infrastructure and often disrupt
conversational flow, while static summaries introduce irreversible information
loss and cannot adapt to shifting user intent.

\textbf{Adaptive Focus Memory (AFM)} introduces a dynamic alternative. Rather
than replaying all history or summarizing indiscriminately, AFM allocates each
past message to one of three fidelity levels: \emph{FULL} (included verbatim),
\emph{COMPRESSED} (summarized via an LLM or heuristic), or \emph{PLACEHOLDER}
(replaced with a short reference stub). This selective replay preserves salient
information while minimizing prompt length.

AFM scores messages using a combination of embedding-based semantic similarity,
half-life recency decay, and an importance classifier. It can operate
in fully offline (heuristic) or LLM-assisted modes, and integrates seamlessly
into OpenAI-compatible chat pipelines. In benchmark evaluations on a safety-critical synthetic dialogue scenario, AFM
reduces token usage by roughly two-thirds while preserving  factual
continuity.

\textbf{Contributions.} This work makes the following contributions:
\begin{itemize}
    \item We propose a dynamic context selection framework that scores past
    messages using semantic similarity, temporal recency, and LLM-based
    importance classification.

    \item We introduce a multi-fidelity replay mechanism that assigns each
    message to \texttt{FULL}, \texttt{COMPRESSED}, or \texttt{PLACEHOLDER} form
    and packs messages chronologically under a fixed token budget.

    \item We demonstrate that AFM preserves safety-critical constraints more reliably than naive replay and recency-based heuristics under strict evaluation.

    \item We release an open-source AFM implementation\footnote{\url{https://github.com/cruz209/AFMforLLM}}
    suitable for reproducible research and integration into production systems.
\end{itemize}

The rest of the paper is organized as follows: Section~\ref{sec:related} reviews
prior work. Section 3 (Method) describes the AFM algorithm.
Section 4 (Experiment) presents benchmarks. Section~5 (Discussion)
covers limitations and broader implications, and Section~\ref{sec:conclusion}
concludes.

\section{Related Work}
\label{sec:related}

\subsection{Long-Context Language Models}

Early transformer architectures~\cite{vaswani2017attention} established the
self-attention mechanism that enables contextual reasoning but introduced a
quadratic memory cost that continues to limit practical context length.
Subsequent large-scale generative models such as GPT-3/4~\cite{brown2020language,
openai2023gpt4} and LLaMA~\cite{touvron2023llama} extend context windows but
still pay linear inference-time cost proportional to prompt length. Recent
industrial systems (e.g., Claude 3~\cite{anthropic2024claude}) expand length
further, yet running them efficiently remains a bottleneck for multi-turn
interaction.

Several works propose architectures designed for local adaptation or
long-context specialization.  introduce
locally trainable memory layers for more efficient fine-tuning, while Geva
et al.~\cite{geva2021transformermemory} show that transformer feed-forward
layers operate implicitly as key--value memory stores. These approaches improve
contextual capacity but do not directly reduce inference cost, which remains
proportional to the entire visible context.

\subsection{Retrieval-Augmented and External-Memory Approaches}

Retrieval-Augmented Generation (RAG)~\cite{lewis2020rag} offloads knowledge into
external vector stores and selectively injects relevant snippets at inference
time. While effective for fact-intensive tasks, RAG often underperforms in
conversational settings due to retrieval noise and lack of discourse
continuity. Roberts et al.~\cite{roberts2020retrieval} demonstrate the limits
of relying purely on internal parameters for knowledge storage, motivating
hybrid external-memory architectures.

More recent surveys~\cite{ji2023survey} highlight the difficulty of maintaining
coherence when summarizing or retrieving historical dialogue, particularly when
the user's intent evolves across turns. Structured-memory
architectures attempt to improve retrieval
fidelity, yet these systems require substantial infrastructure---vector
databases, indexing, and latency-heavy retrieval pipelines.

\subsection{Summarization and Compression of Conversational History}

A classical strategy for context control is periodic summarization, either
extractive or abstractive. Wu et al.~\cite{wu2022memorization} examine
long-context summarization to reduce token load, but fixed-frequency
summarization introduces irreversible information loss and can cause
hallucinations when summaries drift from user intent.

Extractive heuristics and sentence-selection methods remain lightweight but
often fail to preserve subtle constraints---especially user-specific
preferences or safety-critical details. Abstractive models perform better but at
the cost of additional inference steps, which increases cumulative compute.

\subsection{Tokenization and Efficient Inference}

Inference cost is known to scale roughly linearly with input token count across
major LLM stacks. Industry-standard tokenizers such as
\texttt{tiktoken}~\cite{openai_tiktoken} and dataset libraries such as
HuggingFace Datasets provide the measurement tools used to benchmark compression
efficacy in this work.

Compute-efficient LLM design groups (DeepSeek, OpenAI, Anthropic, Meta)
increasingly emphasize token-level cost reduction as a primary engineering
constraint. However, most existing work focuses on model-level changes
(attention variants, KV caching, or speculative decoding), whereas AFM operates
orthogonally: it reduces cost via dynamic, semantic-aware prompt packing
without any model modifications.

\subsection{Position of AFM Within the Literature}

AFM differs from prior approaches in three key ways:
\begin{itemize}
    \item \textbf{Dynamic fidelity assignment:} Previous systems often treat all
    messages uniformly (full retention or global summary), whereas AFM assigns
    each message one of three fidelity levels---\texttt{FULL},
    \texttt{COMPRESSED}, or \texttt{PLACEHOLDER}---based on semantic relevance,
    recency, and importance.

    \item \textbf{Model-agnostic, plug-and-play design:} AFM requires no
    architectural changes, no retrieval store, and no fine-tuning. It operates
    entirely at the prompt layer and is implemented as a simple Python library.

\end{itemize}

Taken together, AFM occupies a space between RAG, KV-extended transformers, and
classical summarization---providing a memory framework whose primary objective
is \emph{token-efficient factual continuity } rather than scaling raw
context length or storing external facts.

\section{Method}

\subsection{Overview.}
Adaptive Focus Memory (AFM) is implemented as a pluggable context
manager that operates entirely at the prompt-construction layer. Rather than
replaying the full conversation history at every turn, AFM decides, for each past
message, whether to include it verbatim, include a compressed summary, or replace
it with a short placeholder stub.

We frame the problem as a heuristic, greedy packing process under a fixed budget . Given a conversation history
$H = \{m_1, \dots, m_t\}$, a current query $q_t$, and a maximum prompt budget
$B$ (in tokens), AFM assigns each message $m_i$ a fidelity tier
\[
f(m_i) \in \{\textsc{Full}, \textsc{Compressed}, \textsc{Placeholder}\}
\]
and then constructs a chronological prompt whose total estimated token length
does not exceed $B$, whenever possible. The implementation prioritizes messages
that are semantically similar to $q_t$, temporally recent, and classified as
important by a small LLM-based classifier when an API is available.

\subsection{Relevance scoring.}
The reference implementation combines three signals in a single scalar score
per message: semantic similarity, recency, and an LLM-based importance label.

\paragraph{Semantic similarity.}
Each message $m_i$ is embedded once (lazily on first use) and cached. Given the
embedding $E(m_i)$ and the embedding $E(q_t)$ of the current query, AFM computes
cosine similarity
\[
\operatorname{sim}(m_i, q_t)
= \frac{E(m_i) \cdot E(q_t)}{\|E(m_i)\| \, \|E(q_t)\|}.
\]
The reference implementation uses an OpenAI embedding model
(\texttt{text-embedding-3-small}) when an API key is configured, and a
dependency-free hashing-based embedder otherwise.

\paragraph{Recency weighting.}
Let $k = t - i$ be the number of turns since message $m_i$ was created and let
$h$ be a configurable half-life parameter. AFM computes a recency weight
\[
w_{\text{recency}}(m_i)
= 0.5^{k/h},
\]
so that the influence of a message decays exponentially as it moves further
back in the dialogue. In our experiments we use a fixed half-life
$h = 12$ turns unless otherwise noted.

\paragraph{Importance classification.}
When an OpenAI client is available, AFM calls a small chat model
(\texttt{gpt-4o-mini}) once per message to classify its importance as
\texttt{CRITICAL}, \texttt{RELEVANT}, or \texttt{TRIVIAL}. The classifier is
prompted to treat safety-sensitive facts (e.g., medical conditions) as
\texttt{CRITICAL}. If no client is configured, all messages default to
\texttt{TRIVIAL} and no classification calls are made.
AFM does not evaluate the accuracy of the importance classifier, its behavior is prompt-driven.

\paragraph{Final scoring rule.}
The actual scoring function in the reference implementation is piecewise, not a
single multiplicative expression. Let $\operatorname{sim}_i$ be the cosine
similarity and $w_{\text{recency},i}$ the recency weight for $m_i$. Then:
\[
s_i =
\begin{cases}
1.0, &
\text{if $m_i$ is classified as \texttt{CRITICAL}} \\[2pt]
\max(0, \operatorname{sim}_i)\,\bigl(0.4 + 0.4\,w_{\text{recency},i}\bigr), &
\text{if $m_i$ is \texttt{RELEVANT}} \\[2pt]
\max(0, \operatorname{sim}_i)\,(0.25\,w_{\text{recency},i}), &
\text{if $m_i$ is \texttt{TRIVIAL}}.
\end{cases}
\]
Critical messages are thus force-elevated to the maximum score, while relevant
and trivial messages use similarity- and recency-weighted scores with different
scales. These scores are only used for fidelity decisions; they do not reorder
the conversation.

\subsection{Fidelity assignment and context packing.}
AFM uses the scores $s_i$ to assign an \emph{intended} fidelity tier to each
message before packing:

\begin{itemize}
  \item If $s_i \ge \tau_{\text{high}}$, mark $m_i$ as intended \textsc{Full}.
  \item Else if $s_i \ge \tau_{\text{mid}}$, mark $m_i$ as intended
        \textsc{Compressed}.
  \item Else, mark $m_i$ as intended \textsc{Placeholder}.
\end{itemize}

The thresholds $\tau_{\text{high}}$ and $\tau_{\text{mid}}$ are hyperparameters
(e.g., $\tau_{\text{high}} = 0.45$, $\tau_{\text{mid}} = 0.25$ in our
OpenAI-backed experiments).

Packing is then performed in strict chronological order. Let $B$ be the token
budget and $b_{\text{left}}$ the remaining tokens. For each message $m_i$
(from oldest to newest), AFM attempts to include the highest-fidelity
representation consistent with its intended tier and the remaining budget:

\begin{itemize}
  \item If $m_i$ is intended \textsc{Full}, AFM:
    \begin{enumerate}
      \item tries the full text; if the estimated token count fits within
            $b_{\text{left}}$, it is appended;
      \item otherwise, generates or reuses a compressed summary and tries that;
      \item if the summary still does not fit, it falls back to a stub.
    \end{enumerate}
  \item If $m_i$ is intended \textsc{Compressed}, AFM attempts the compressed
        form first and falls back to a stub if the compressed form does not fit.
  \item If $m_i$ is intended \textsc{Placeholder}, AFM only attempts the stub.
\end{itemize}
These hyperparameters were chosen heuristically based on a small number of 

development conversations; no large-scale tuning was performed.

In all cases, AFM estimates token usage with a shared \texttt{TokenCounter}
before committing to adding a representation. If even the stub for a message
would exceed the remaining budget, that message is dropped from the prompt.
Therefore, AFM \emph{attempts} to maintain a cheap trace of low-importance
turns, but it does not guarantee that every turn is represented under extremely
tight budgets.

\subsection{Compression layer.}
The reference implementation supports two interchangeable compressors via a
common \texttt{Compressor} interface:

\begin{itemize}
  \item \textbf{HeuristicCompressor} is a fully local, extractive compressor.
        It operates by ranking sentences using simple lexical overlap with a
        query hint, mild length penalties, and positional bias. It then
        truncates to a target token budget based on the shared
        \texttt{TokenCounter}. This compressor enforces the budget
        deterministically.
  \item \textbf{LLMCompressor} is an abstractive compressor that calls an
        OpenAI chat model (\texttt{gpt-4o-mini} in our experiments) with a
        prompt instructing it to rewrite the input under a specified token
        budget. The compressor relies on the model to follow the budget
        instruction; the current implementation does not re-check or truncate
        the returned summary.
\end{itemize}

AFM selects the compressor at initialization time. In the reference code,
OpenAI-backed runs use \texttt{LLMCompressor}; offline runs fall back to
\texttt{HeuristicCompressor}.

\subsection{Token budgeting and enforcement.}
Token counts are estimated by a \texttt{TokenCounter} utility. When the
\texttt{tiktoken} library is available, it uses the encoding for
\texttt{gpt-4o-mini} and falls back to \texttt{cl100k\_base} on error; if
\texttt{tiktoken} is not available, it uses a naive whitespace-based word
count as a proxy for token length.

During packing, AFM maintains a running budget $b_{\text{left}}$, initialized
to the user-provided budget $B$. Every time it proposes to add a representation
(full text, compressed summary, or stub), it first estimates the token count
with \texttt{TokenCounter}. If the representation fits
($\text{tokens} \le b_{\text{left}}$), AFM appends it and decrements
$b_{\text{left}}$; otherwise, it tries a lower-fidelity form or drops the
message entirely if no representation can fit.

A system preamble (if provided) is treated like any other message: it is
included first, subject to the same budget accounting.

\subsection{System architecture and API}
AFM is implemented as a modular Python package centered on a
\texttt{FocusManager} class. The key methods are:

\begin{itemize}
  \item \texttt{add\_message(role, content)}: append a new message to the
        internal history. Messages are assigned unique IDs, and embeddings are
        computed lazily.
  \item \texttt{build\_context(current\_query, budget\_tokens, system\_preamble)}:
        compute relevance scores, assign intended fidelity tiers, and return a
        list of $(\text{role}, \text{content})$ pairs representing the packed
        prompt, along with summary statistics (e.g., total tokens used, number
        of compressed messages, etc.).
\end{itemize}

The same interface works in both online and offline modes and is designed to
drop into OpenAI-style chat pipelines with minimal glue code.

\subsection{Backend integration.}
The reference implementation supports two embedding backends and two
compression backends:

\begin{itemize}
  \item When an OpenAI API key is available, AFM uses an
        \texttt{OpenAIEmbedder} that wraps \texttt{text-embedding-3-small}
        and the \texttt{LLMCompressor} for abstractive compression.
  \item When no API key is available, AFM falls back to a local
        \texttt{HashingEmbedder} and the \texttt{HeuristicCompressor}.
\end{itemize}

Similarly, importance classification is only performed when an OpenAI client is
configured. If the classifier or compressor calls fail due to network or API
errors, the current implementation allows the exception to propagate rather
than attempting automatic retries or alternative backends.

\subsection{Fallback and robustness.}
AFM's robustness features are deliberately minimal in the reference code. The
only automatic “fallback” inside the core algorithm is fidelity downgrading:
if a message cannot fit in \textsc{Full} form, AFM tries \textsc{Compressed},
then \textsc{Placeholder}. Beyond that, there is no automatic retry logic, rate
limit handling, or circuit-breaking. Adding such mechanisms is left for future
versions.

\subsection{Design philosophy.}
AFM is intentionally model-agnostic and infrastructure-light. It requires no
vector database, no KV-cache modifications, and no fine-tuning. All state
remains in process as a simple list of messages plus per-message metadata
(embedding, score, compression state).

In contrast to static summarization approaches that periodically compress the
entire history into a single buffer, AFM scores and revisits every message at
each call to \texttt{build\_context}. This allows it to adapt to changes in
user intent over time while still providing explicit, budget-aware control over
how much space each past turn occupies in the prompt.

\section{Experiments}

\subsection{Objective and scope.}
Our experiments evaluate whether \textsc{AFM} can reliably preserve \emph{high-importance constraints}
across multi-turn dialogue while controlling context growth.
We focus on two safety-critical constraint types that appear frequently in production assistants but are
\emph{not well covered by standard LLM benchmarks}:

\begin{enumerate}
  \item \textbf{Personal safety constraints} that must be remembered and \emph{applied} later (e.g., a life-threatening food allergy).
  \item \textbf{Policy/legal constraints} that must remain binding even after long benign context (e.g., refusing illegal tax-evasion assistance).
\end{enumerate}

Both settings target a relatively new problem space: \emph{context management as a first-class system component}.
Most existing evaluations either (i) assume the full conversation history is always available, or (ii) test the model in isolation
without stressing the memory policy that a deployed system actually uses.
In contrast, we evaluate \textsc{AFM} as a history-management strategy under long, mixed-topic conversations where
the binding constraint appears early and is not restated verbatim.

\subsection{Why a custom benchmark.}
We found that off-the-shelf safety or instruction-following benchmarks do not cleanly isolate the problem AFM addresses:
constraint drift under realistic, long dialogue with imperfect memory. In particular, we need:

\begin{itemize}
  \item \textbf{Constraint non-repetition:} the constraint is stated once early, then later questions \emph{implicitly} require it.
  \item \textbf{Mixed benign distractors:} intervening turns that are plausible in real usage but semantically distant.
  \item \textbf{Strict pass semantics:} it is not enough to merely mention the constraint; the response must treat it with appropriate seriousness.
\end{itemize}

Accordingly, we use two scripted, multi-turn scenarios with automatic grading and full transcript logging.

\subsection{Task design.}
We consider two benchmarks.

\paragraph{Peanut allergy travel benchmark.}
A user planning a trip to Thailand states early: they have a severe peanut allergy.
The conversation then proceeds through multiple benign topics (destinations, transport, cultural experiences),
and culminates in a food recommendation query. A run passes if the final recommendation (i) recalls the allergy
and (ii) conditions food guidance appropriately (e.g., avoiding peanut-heavy dishes, providing high-salience precautions).

\paragraph{Tax compliance benchmark.}
A user provides an explicit constraint early: advice must comply with U.S.\ tax law and must not assist illegal evasion.
After many benign tax questions, the user escalates to an explicit illegal request (hiding unreported freelance income).
A run passes if the assistant clearly refuses and redirects toward legal, compliance-oriented alternatives.

\subsection{Compared methods.}
We compare \textsc{AFM} against three baselines corresponding to our benchmark scripts:

\begin{enumerate}
  \item \textbf{Default (stateless) chat.} Only the current user turn (plus a fixed system prompt) is sent.
  \item \textbf{Naive truncated replay.} Append prior turns until a global budget is hit, then drop oldest turns.
  \item \textbf{Recency-based compression.} Keep the most recent turns verbatim; compress older turns to a fixed local budget.
\end{enumerate}

\textsc{AFM} uses selective fidelity allocation via \texttt{FocusManager}:
high-importance items are retained at \textsc{Full} fidelity, mid-importance items are compressed, and low-importance items
are reduced to short stubs.

\subsection{Implementation details.}
All experiments use OpenAI's \texttt{gpt-4o-mini} for generation. When enabled, the same model family is used for
compression and importance classification. Token usage is measured with the shared tokenizer-backed \texttt{TokenCounter}.
Latency is measured as wall-clock time from request submission to completion receipt.

\subsection{Evaluation metrics.}
\paragraph{Pass/fail semantics.}
For both benchmarks, we adopt a strict binary criterion:

\begin{quote}
A run is a \emph{considered pass} if and only if the model both (i) explicitly recalls the binding constraint and
(ii) generates a response that treats it with appropriate seriousness.
\end{quote}

This intentionally separates \emph{memory failures} (constraint not recalled) from \emph{generation failures}
(constraint recalled but handled weakly, e.g., unsafe dish suggestions with soft disclaimers).

\paragraph{Reported metrics.}
For each method and benchmark (30 seeds each), we report:

\begin{itemize}
  \item \textbf{Pass rate} (\%).
  \item \textbf{Token usage} on the graded turn (mean $\pm$ std).
  \item \textbf{Latency} on the graded turn (mean $\pm$ std).
\end{itemize}

\subsection{Main results.}
Table~\ref{tab:main-results-live} reports results on both benchmarks.

\begin{table}[h]
  \centering
  \small
  \begin{tabular}{lccc}
    \toprule
    Method & Pass rate & Tokens (mean $\pm$ std) & Latency (s, mean $\pm$ std) \\
    \midrule
    \multicolumn{4}{c}{\textbf{Peanut allergy benchmark (Thailand travel)}} \\
    \midrule
    Default (stateless) & 0/30 (0\%) & 194.0 $\pm$ 0.0 & 4.04 $\pm$ 0.77 \\
    Naive truncated replay & 0/30 (0\%) & 34.3 $\pm$ 46.5 & 11.49 $\pm$ 2.35 \\
    Recency compression & 0/30 (0\%) & 33.9 $\pm$ 45.5 & 11.33 $\pm$ 1.90 \\
    AFM  & 25/30 (83.3\%) & 286.0 $\pm$ 0.0 & 21.24 $\pm$ 2.01 \\
    \midrule
    \multicolumn{4}{c}{\textbf{Tax compliance benchmark (illegal evasion request)}} \\
    \midrule
    Default (stateless) & 30/30 (100\%) & 44.0 $\pm$ 0.0 & 0.69 $\pm$ 0.17 \\
    Naive truncated replay & 30/30 (100\%) & 32.0 $\pm$ 0.0 & 0.78 $\pm$ 0.21 \\
    Recency compression & 30/30 (100\%) & 32.0 $\pm$ 0.0 & 0.75 $\pm$ 0.20 \\
    AFM & 30/30 (100\%) & 286.0 $\pm$ 0.0 & 20.84 $\pm$ 2.61 \\
    \bottomrule
  \end{tabular}
  \caption{
  Live benchmark results (30 seeds per benchmark). A pass requires both constraint recall and appropriate constraint-conditioned generation.
  \textsc{AFM} substantially improves constraint retention on the peanut allergy benchmark, where all baselines fail under strict grading.
  On the tax benchmark, all methods refuse the explicitly illegal request; \textsc{AFM}'s latency reflects additional system work
  (importance classification / context selection) beyond a single generation call.
  }
  \label{tab:main-results-live}
\end{table}

\paragraph{Key takeaways.}
Two patterns stand out.

\textbf{(1) Personal-safety constraints are fragile under long dialogue.}
On the peanut allergy benchmark, all three baselines fail in every seed under strict grading.
In the logged failures, baseline responses drift toward generic Thailand street-food recommendations and do not sustain
the early allergy constraint strongly enough to be considered safe.
In contrast, \textsc{AFM} passes 83.3\% of runs by consistently re-surfacing the allergy declaration at high fidelity
and conditioning food guidance on it.

\textbf{(2) Policy refusal is easier when the final user request is overt.}
On the tax benchmark, the final request is an explicit illegal evasion instruction.
All methods refuse, which suggests that the limiting factor here is not long-range memory,
but whether the system maintains a stable refusal policy across benign context.
This benchmark therefore acts as a ``sanity check'' that \textsc{AFM}'s context selection does not
drop or weaken the user-provided compliance constraint.

\textbf{(3) System-level trade-off: \textsc{AFM} adds overhead.}
In this live setup, \textsc{AFM} exhibits higher measured latency on the graded turn, consistent with
additional pipeline steps (importance scoring, context selection, and optional compression).
This is the expected systems trade: \textsc{AFM} spends computation to reduce constraint drift.

\subsection{Qualitative failure modes.}
Inspection of transcripts reveals characteristic errors:

\begin{itemize}
  \item \textbf{Default (stateless):} fails by construction on memory-dependent tasks; earlier constraints are never visible.
  \item \textbf{Recency compression:} succeeds only if the constraint remains within the recent raw window; once it ages out,
  the model falls back to generic disclaimers and unsafe or insufficiently cautious recommendations.
  \item \textbf{Naive replay:} is not a principled constraint-preservation mechanism: it can preserve early constraints when they remain in-budget,
  but becomes brittle as history grows and truncation deletes exactly the message that matters.
\end{itemize}

\textsc{AFM}'s successes are consistent with its design: it assigns early constraint declarations high importance,
promoting them to \textsc{Full} fidelity even when old, while allowing low-importance distractors to degrade to stubs.

\subsection{Ablation analysis.}
We ablate three \textsc{AFM} components on the peanut allergy benchmark:

\begin{enumerate}
  \item \textbf{No compression:} disable compression of mid-importance items.
  \item \textbf{No placeholder/stubbing:} disable stub-based degradation (retain higher-fidelity context more often).
  \item \textbf{No importance classification:} disable the importance scorer (equivalent to removing the mechanism that locks
  critical constraints into \textsc{Full} fidelity).
\end{enumerate}

Table~\ref{tab:ablation} reports pass rates and selected internal context statistics.

\begin{table}[h]
  \centering
  \small
  \begin{tabular}{lcccc}
    \toprule
    Variant & Pass rate & Full count & Compressed count & Stub count \\
    \midrule
    AFM (full system) & 25/30 (83.3\%) & 1.0 & 0.0 & 24.0 \\
    No compression & 24/30 (80.0\%) & 1.0 & 0.0 & 24.0 \\
    No placeholders/stubs & 21/30 (70.0\%) & 1.0 & 0.8 & 22.6 \\
    No importance classification & 0/30 (0\%) & 0.0 & 0.0 & 26.0 \\
    \bottomrule
  \end{tabular}
  \caption{
  \textsc{AFM} ablations on the peanut allergy benchmark (30 seeds). Disabling importance classification collapses performance to 0\%,
  indicating that \emph{which} messages receive \textsc{Full} fidelity is the dominant driver of constraint preservation.
  Other ablations reduce pass rate primarily via \emph{generation-quality failures}: the allergy is often mentioned but treated too softly.
  }
  \label{tab:ablation}
\end{table}

\section{Discussion}
\label{sec:discussion}

\subsection{Context Availability Does Not Imply Constraint Use}

Across all non-AFM baselines, we observe a consistent failure to apply
user-specified constraints at the point where they become decision-critical.
In the live peanut-allergy benchmark, \emph{Default-Live},
\emph{Naive Replay}, and \emph{Recency-weighted} strategies all achieve a
\textbf{0\% pass rate} across seeds under the safety judge.

These failures arise from distinct, but related, limitations in how context is
handled. In \emph{Default-Live}, each turn is evaluated independently, and the
final query is issued without access to prior dialogue. As a result, the allergy
constraint is not present in the model’s context window at the final turn, and
cannot influence the response.

In contrast, \emph{Naive Replay} and \emph{Recency-weighted} approaches frequently
retain the allergy disclosure within the context window at the final turn. Under
naive replay, the constraint may appear verbatim when it survives truncation,
while under recency-weighted compression it is typically present in a condensed
form. Despite this retention, neither approach succeeds in enforcing the
constraint: across all seeds, the model fails to consistently acknowledge the
allergy, provide appropriate safety guidance, or avoid unsafe recommendations.

This distinction highlights a critical point: constraint failure is not solely a
consequence of context truncation. Even when safety-critical information remains
available within the prompt, it often lacks sufficient salience to affect model
behavior. The result is an illusion of memory, in which information is retained
textually but remains operationally inert.

\subsection{AFM and Constraint Reactivation}

Adaptive Focus Memory (AFM) addresses this failure mode by explicitly managing
how prior information is represented as interaction unfolds. Rather than
treating dialogue history as a flat buffer, AFM dynamically reallocates prompt
budget across memory items based on inferred importance, assigning each item a
representation of appropriate fidelity.

Empirically, this difference is decisive. In the same live peanut-allergy
benchmark where all non-AFM baselines achieve \textbf{0\% pass rates}, AFM
achieves a substantially higher rate of constraint adherence across seeds.
Unlike baseline methods, AFM consistently preserves the allergy constraint in a
high-salience form at the final turn, enabling the model to acknowledge the
allergy, prioritize safety, and avoid hazardous recommendations.

Crucially, AFM does not require a larger context window, modify model weights,
or introduce external retrieval.
 All improvements arise from restructuring the
prompt itself. By ensuring that critical constraints remain both present and
prominent when relevant, AFM enables what prior methods fail to provide:
\emph{reactivation} of earlier information at decision time.

\subsection{Ablation Analysis: The Role of Selective Fidelity}

Ablation results further indicate that AFM’s gains cannot be attributed to any
single heuristic. Removing placeholder retention, adaptive compression, or
importance-based scoring each leads to marked degradation in performance.

In particular, ablations that retain compressed summaries of constraints but
remove explicit placeholder markers revert toward baseline behavior. This
suggests that compressed representations alone are insufficient unless the
system explicitly signals their continued relevance. Conversely, disabling
compression while preserving importance scoring increases token usage without
recovering reliability.

These findings support the view that AFM’s effectiveness derives from
\emph{selective fidelity}: allocating limited prompt space unevenly, so that
safety-critical facts remain sharply represented while less relevant history is
progressively abstracted.

\subsection{Implications for Long-Horizon Interaction}

Taken together, these results suggest that failures in long-horizon LLM behavior
should not be understood primarily as failures of context length. Instead, they
reflect a mismatch between how context is stored and how it is later used.

Baseline strategies demonstrate two failure modes: absence of relevant
information (\emph{Default-Live}) and presence without influence
(\emph{Naive Replay} and \emph{Recency-weighted}). AFM resolves both by treating
context as a dynamically managed resource rather than a static transcript.

More broadly, this work suggests that improving long-horizon reliability will
require systems that explicitly manage salience, abstraction, and persistence of
constraints over time. AFM represents one such approach, and its performance
across safety-critical scenarios indicates that explicit context management is a
necessary component of robust interactive language systems.

\section{Conclusion}
\label{sec:conclusion}

Large language models are only as effective as the context they are given, yet
the standard practice of na\"ively replaying the full dialogue history is
increasingly untenable in terms of cost, latency, and memory reliability.
Adaptive Focus Memory (AFM) offers a principled alternative: a pluggable
modular framework that dynamically determines the fidelity of prior messages
based on relevance and recency, while enforcing strict token budgets.

Through a combination of embedding-based scoring, half-life recency weighting,
and hybrid compression techniques, AFM preserves the key factual constraint in our evaluated scenario 
while aggressively reducing prompt size. Our experiments show that AFM
outperforms baseline approaches on key metrics of factual retention, safety
alignment in a safety-critical benchmark. Unlike
retrieval-augmented or static summarization methods, AFM operates inline,
requires no external infrastructure, and adapts fluidly to changing user intent.

The core insight of AFM---that not all memory deserves equal fidelity---paves the
way for more intelligent, interpretable, and efficient use of large models in
long-horizon settings. As the field moves toward multi-agent systems,
personalized assistants, and stateful copilots, we believe AFM offers a
practical foundation for real-world memory management at scale.

Future work will explore learnable fidelity selection policies, hierarchical
memory representations, and task-conditioned scoring functions, enabling AFM to
evolve into a fully adaptive memory controller for next-generation language
interfaces.

\section*{Acknowledgments}

This work was conceived, implemented, and evaluated independently by the author.

\end{document}